\def\year{2020}\relax
\documentclass[letterpaper]{article} 
\usepackage{aaai20}  
\usepackage{times}  
\usepackage{helvet} 
\usepackage{courier}  
\usepackage[hyphens]{url}  
\usepackage{graphicx} 
\usepackage{graphicx}  
\usepackage[skip=1pt,labelformat=simple]{subcaption}
\DeclareCaptionFont{tiny}{\tiny}
\newcommand{\citet}[1]{\citeauthor{#1}\shortcite{#1}}
\newcommand{\citep}{\cite}

\title{Idle Time Optimization for Target Assignment and Path Finding\\in Sortation Centers
}
\author{Ngai Meng Kou, Cheng Peng\\ Cainiao Smart Logistics Network\\ \{ngaimeng.knm, junpeng.pc\}@alibaba-inc.com \And Hang Ma \\ Simon Fraser University\\ hangma@sfu.ca \And T. K. Satish Kumar, Sven Koenig\\ University of Southern California
\\ tkskwork@gmail.com, skoenig@usc.edu
}

\usepackage{amsfonts} 
\usepackage{amsthm}
\theoremstyle{plain}
\newtheorem{thm}{Theorem}

\providecommand{\route}[1]{\langle {#1} \rangle}
\providecommand{\startt}[1]{T^s_{#1}}
\providecommand{\goalt}[1]{T^g_{#1}}
\providecommand{\station}{s}
\providecommand{\start}{\mathfrak{s}}
\providecommand{\goal}{\mathfrak{g}}

\usepackage[linesnumbered,vlined,ruled]{algorithm2e}
\makeatletter
\newcommand\notsotiny{\@setfontsize\notsotiny{7}{8}}
\makeatother
\SetKwProg{Fn}{Function}{}{}
\SetAlFnt{\scriptsize}
\SetAlCapFnt{\normalfont\scriptsize}
\SetAlCapNameFnt{\scriptsize}
\SetArgSty{textrm}
\SetInd{0.1em}{0.5em}
\SetCommentSty{emph}
\usepackage{siunitx}

\begin{document}
	
	\maketitle
	
	\begin{abstract}
		In this paper, we study the one-shot and lifelong versions of the Target Assignment and Path Finding problem in automated sortation centers, where each agent needs to constantly assign itself a sorting station, move to its assigned station without colliding with obstacles or  other agents, wait in the queue of that station to obtain a parcel for delivery, and then deliver the parcel to a sorting bin. The throughput of such centers is largely determined by the total idle time of all stations since their queues can frequently become empty. To address this problem, we first formalize and study the one-shot version that assigns stations to a set of agents and finds collision-free paths for the agents to their assigned stations. We present efficient algorithms for this task based on a novel min-cost max-flow formulation that minimizes the total idle time of all stations in a fixed time window. We then demonstrate how our algorithms for solving the one-shot problem can be applied to solving the lifelong problem as well. Experimentally, we believe to be the first researchers to consider real-world automated sortation centers using an industrial simulator with realistic data and a kinodynamic model of real robots. On this simulator, we showcase the benefits of our algorithms by demonstrating their efficiency and effectiveness for up to 350 agents.
	\end{abstract}
	
	\section{Introduction}
	With the increasing popularity of e-commerce and recent progress in AI and robotics research, hundreds of warehouse robots have been employed to sort express parcels in modern automated warehouses and sortation facilities. For example, one such automated sortation center has contributed to a new one-day sales record of \$30.8 billion during the Singles Day Online Shopping Festival (November 11, 2018), that also resulted in the delivery of over one billion express parcels throughout China within one week \cite{Alizila}.
	
\begin{figure}[t]
	\centering%
	\begin{subfigure}[t]{.54\columnwidth}
		\centering
		\includegraphics[width=\columnwidth]{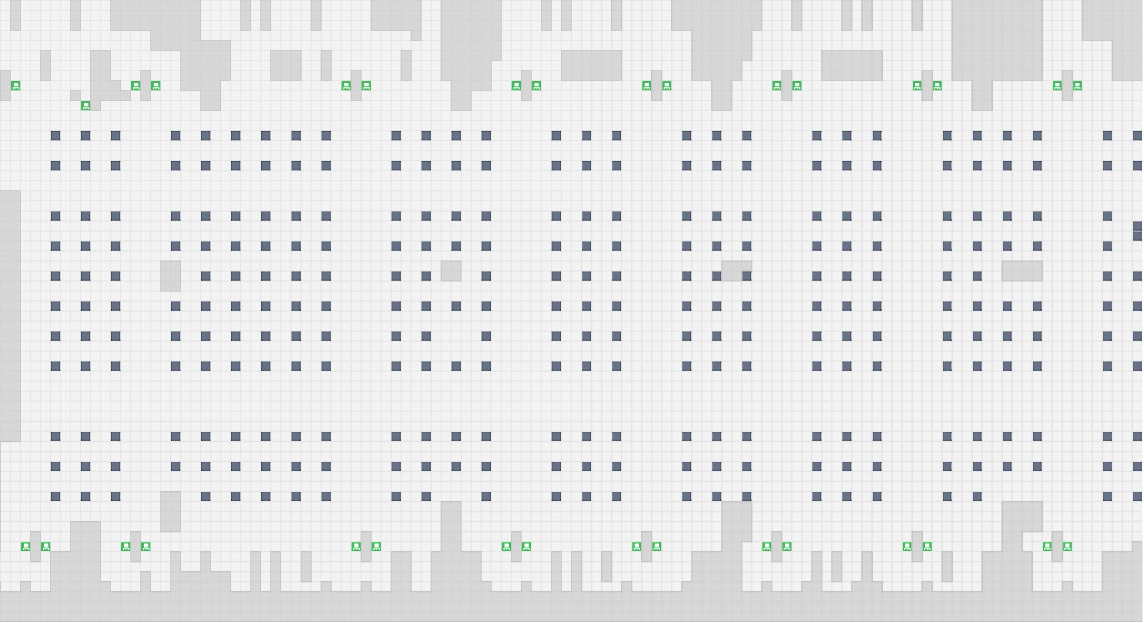}
		\caption{The 2D grid layout of an automated sortation center. Green cells represent sorting stations. Gray cells represent obstacles. Blue cells represent sorting bins.}
		\label{fig:sortation_map_real}
	\end{subfigure}\hfill%
	\begin{subfigure}[t]{.44\columnwidth}
		\centering		
		\includegraphics[width=\columnwidth]{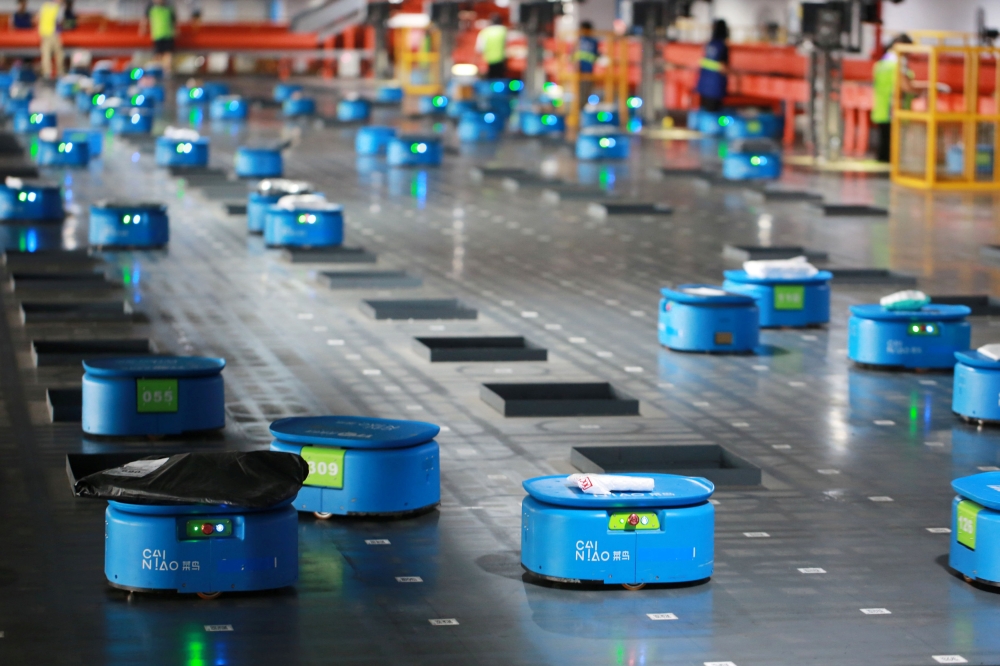}
		\caption{Robots delivering parcels to sorting bins \protect\cite{Reuters}.}
		\label{fig:robot_sorting_system}
	\end{subfigure}%
	\caption{Example of an automated sortation center.}%
	\label{fig:sortation}%
\end{figure}

	Figure \ref{fig:sortation_map_real} shows the layout of a typical sortation center. In such a center, each agent has to assign itself a sorting station, move to its assigned station without colliding with obstacles or other agents, and then wait in the queue of that station to obtain a parcel for delivery. The earliest agent in the queue obtains a parcel from the station and delivers it to its destination, a sorting bin located in a designated cell. Typically, a sorting bin is associated with the shipping addresses of a specific ZIP code. Figure \ref{fig:robot_sorting_system} shows agents in action in such a center. At each sorting station, a human worker or a machine scans the barcode of a parcel and determines its sorting bin. After an agent from the queue of the sorting station successfully delivers the parcel, it can retask itself by waiting in another queue. In general, there are many more parcels to be delivered than agents available for delivery. Therefore, empty queues, i.e., idle times of sorting stations, are the bottleneck for the throughput of the center. In this paper, we aim to minimize the total idle time of such a sortation center.
	
%

	\subsection{Related Work on Task and Path Planning}
	
	\noindent\textbf{One-Shot Path Planning:} Past research on multi-agent path-planning problems has mostly centered around \emph{Multi-Agent Path Finding (MAPF)} \cite{MaAIMATTERS17,SternSOCS19}. MAPF aims to find collision-free paths for a given set of agents from their current locations to given targets. In general, it is NP-hard to solve optimally for minimizing flowtime (the sum of the arrival times of all agents at their targets) and NP-hard to approximate within any constant factor less than 4/3 for minimizing makespan (the maximum of the arrival times of all agents at their targets) \cite{surynek2010optimization,YuLav13AAAI,MaAAAI16}. It can be solved via reductions to other well-studied combinatorial problems \cite{YuLav13ICRA,Surynek15,erdem2013general} or by specialized rule-based, search-based, and hybrid algorithms \cite{PushAndSwap,Wang11,PushAndRotate,DBLP:journals/ai/SharonSGF13,MStar,DBLP:journals/ai/SharonSFS15,LiIJCAI19,LiICAPS19,LamBHS19,GangeHS19}. MAPF is insufficient for modeling sortation centers since it assumes that a target is assigned to each agent a priori and thus does not address the problem of assigning targets to agents, i.e., the target-assignment problem.
	
	\noindent\textbf{One-Shot Task and Path Planning:} Classical algorithms for assignment problems \cite{Kuhn1955,gross1959} can be used to assign targets to agents so that the flowtime or makespan is minimized. Recent research has also considered One-to-One Multi-Team \emph{Target Assignment and Path Finding (TAPF)} \cite{MaAAMAS16}, where agents are partitioned into teams. Each team is given the same number of targets as there are agents in the team. One-to-One Multi-Team TAPF aims to assign targets to agents and plan collision-free paths for the agents to their assigned targets in a way such that each agent moves to exactly one of the targets given to its team, all targets are visited, and the makespan is minimized. A target of a team can be assigned to any agent in the team. The agents in the same team are thus \emph{anonymous}, i.e., interchangeable. One-to-One Multi-Team TAPF is NP-hard to solve optimally if more than one team exists. One-to-One One-Team TAPF, also called \emph{Anonymous MAPF}, is the special case where only one team of (anonymous) agents exists. It can be solved optimally for makespan minimization in polynomial time via a max-flow algorithm \cite{YuLav13STAR}. 
	
	\noindent\textbf{Lifelong Task and Path Planning:} All problems mentioned above are \emph{one-shot} problems since each agent visits exactly one target or executes exactly one task, and then stops there forever. Recent research has also considered the \emph{lifelong} task- and path-planning problem \emph{Multi-Agent Pickup and Delivery (MAPD)} \cite{MaAAMAS17,MaAAAI19b}, where a given team of agents has to attend to a stream of tasks that appear at unknown times and are each characterized by a pickup and a delivery location. MAPD aims to repeatedly solves the target-assignment and path-finding problem, i.e., assigns tasks to agents and finds collision-free paths for all agents from their current locations to the pickup locations and then to the delivery locations of their assigned tasks, whenever there are unexecuted tasks and agents available for executing them. The effectiveness of a MAPD solution is measured by its makespan or service time (the average time to finish each task after it appears). Other research has considered offline versions of MAPD, where all tasks are known a priori \cite{GTAPF,LiuAAMAS19}. Recent research has also considered Online MAPF \cite{vsvancara2019online}, where agents with preassigned targets appear at unknown times.
	
	\subsection{Contributions}
	
	The lifelong version of the task- and path-planning problem in a sortation center shares similarities with TAPF and MAPD since agents need to repeatedly assign sorting stations to themselves and move to their assigned stations to obtain parcels for delivery. However, none of the existing research aims to minimize the idle time of the sorting stations. In this paper, we thus study the task- and path-planning problem in an automated sortation center as a lifelong extension of TAPF.
	
	As our first contribution, we formalize and study a one-shot version of TAPF that assigns sorting stations to a set of agents and finds collision-free paths for them to their assigned stations. This one-shot problem addresses the throughput consideration for automated sortation centers. It also occurs at the crux of other real-world applications, including automated aircraft towing systems \cite{airporttug16}, that assign runways to and find paths for autonomous vehicles that tow airplanes from terminal gates to runways for takeoff.
	
	As our second contribution, we present two efficient algorithms for solving the one-shot version of TAPF based on a novel min-cost max-flow framework, called the \emph{Idle Time Optimization (ITO)} flow framework. The first algorithm uses the ITO flow framework to first assign stations to agents, using their estimated arrival time at every station as path cost, so that the (estimated) total idle time is minimized. It then uses a MAPF algorithm to find collision-free paths for the agents to the stations. The second algorithm combines the ITO flow framework and the MAPF flow framework \cite{YuLav13STAR}. The resulting framework, called the \emph{Path Finding with ITO (PITO)} flow framework, simultaneously assigns stations to and finds collision-free paths for the agents so that the (actual) total idle time is minimized.
	
	As our third contribution, we demonstrate how our algorithms for the one-shot version of TAPF can be applied to the lifelong version of TAPF. We also develop intelligent baseline algorithms for the lifelong version of TAPF, which directly use existing target-assignment and path-finding algorithms. All our algorithms are applicable to realistic automated sortation centers. In general, our ITO- and PITO-based algorithms outperform the baseline algorithms. We believe to be the first researchers to consider real-world automated sortation centers using an industrial simulator with realistic data and a kinodynamic model of real robots. On this simulator, we showcase the benefits of our algorithms by demonstrating their efficiency and effectiveness for up to 350 agents.
	
	\section{One-Shot TAPF}
	
	In this section, we formalize the one-shot version of TAPF, \emph{One-Shot TAPF}. One-Shot TAPF assigns stations to a given set of agents and finds collision-free paths for them to their assigned stations. We are given (1) a connected undirected graph $G = (V, E)$ whose vertices $V$ correspond to locations and whose edges $E$ correspond to connections between the locations that the agents can traverse, (2) a set of $N$ stations $\{\station_j|j=1,\ldots, N\}$, where each station $\station_j$ is associated with a unique target $\goal_j\in V$ where an agent can obtain a parcel, and (3) a set of $M$ agents $\{a_i|i=1,\ldots,M\}$, where each agent $a_i$ enters the environment at its given \emph{start time step} $\startt i$ in its given start location $\mathfrak{s}_i\in V$.
	An {\em assignment} $\sigma$ of stations to agents maps each agent $a_i$ to one station $\station_j = \sigma(a_i)$, while a station might be assigned to any number of agents.
	
	Let $\pi_i(t)$ be the location of agent $a_i$ at time step $t$. A \emph{path} $\pi_i$ for agent $a_i$ is a sequence of locations that satisfies the following conditions: The agent (1) starts in its start location $\start_i$ at its start time step $\startt i$, i.e., $\pi_i(\startt i) = \start_i$; (2) always either moves to an adjacent location or waits in its current location between two consecutive time steps, i.e., $(\pi_i(t), \pi_i(t+1)) \in E$ or $\pi_i(t+1) = \pi_i(t)$ for all time steps $t$; and (3) ends in the target associated with its assigned station at its \emph{arrival time step} $\goalt i$, i.e., $\pi_i(\goalt i) = \goal_j$ such that $\station_j = \sigma(a_i)$. Agents have to avoid collisions with each other: (1) Two agents cannot be in the same location at the same time step (vertex collision); and (2) two agents cannot traverse the same edge in opposite directions at the same time step (edge collision).
	
	We assume that the \emph{processing time} for a human worker or a machine at each station to task an agent with a parcel is $T$ time steps. Therefore, for a given time window $[0,KT)$, the operation time of each station $s_j$ can be divided into $K$ \emph{working slots}, i.e., $K$ sequential periods of $T$ time steps each: $[0,T), [T,2T), \ldots, [kT,(k+1)T), \ldots, [(K-1)T, KT)$.
	Each station admits an agent only at time steps $kT$ for $k=0,\ldots, K-1$: If agent $a_i$ is in target $\goal_j$ at time step $kT$, it is removed from graph $G$ at time step $kT+1$ and occupies working slot $[kT,(k+1)T)$ of station $s_j$. Thus, no idle time occurs during working slot $[kT,(k+1)T)$ for the station.
	These assumptions are simplifying since, in reality, an agent in target $\goal_j$ at time step $t$ other than $0, T, \ldots, (K-1)T$ could occupy station $s_j$ during time steps $[t,t+T)$.
	We use them for better exposition of the workflow of the stations in our problem, even though our algorithms could easily be generalized beyond them.
	
	The \emph{total idle time} is $T$ times the sum of the number of unoccupied working slots over all stations. To allow for flexibility in changing the number of \emph{active} agents, a NULL station that does not contribute to the total idle time is introduced. Agents that are assigned the NULL station are not active within the given time window. Inactive agents do not have paths. We assume that they are removed from the environment and thus do not block other agents since, in reality, sortation centers often exhibit well-formedness \cite{CapVK15}, where agents can stay at their start locations without blocking other agents. A solution to One-Shot TAPF consists of an assignment of stations to all active agents and collision-free paths for them to their assigned stations. An optimal solution to One-Shot TAPF is a solution that minimizes the total idle time within a given time window.
	
	\begin{figure}[t]
		\centering
		\includegraphics[width=.38\columnwidth]{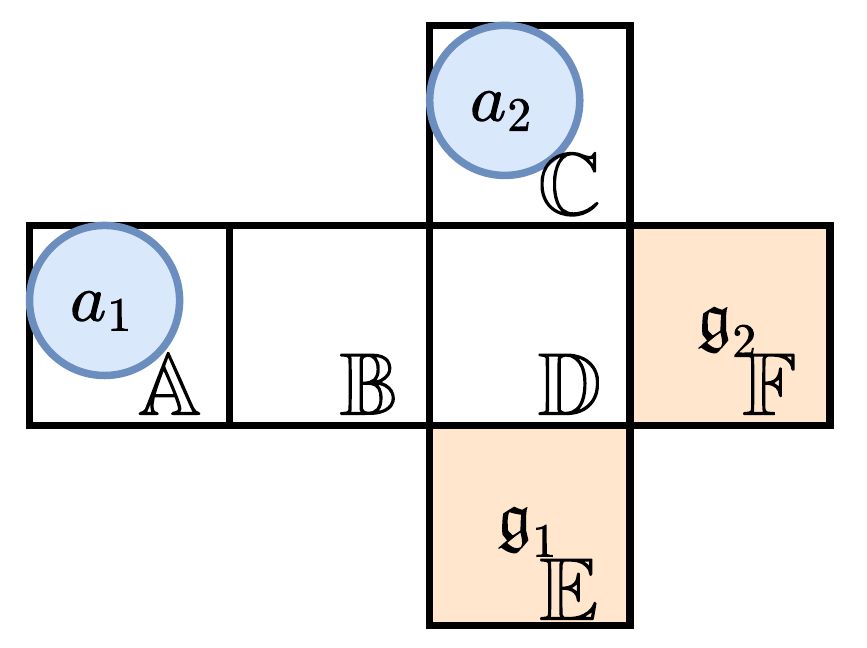}
		\caption{Running example for One-Shot TAPF. Blue circles represent agents. Orange cells represent targets.}
		\label{fig:MAPF}
	\end{figure}
	
	Figure \ref{fig:MAPF} shows a One-Shot TAPF instance on a four-neighbor 2D grid that is used as a running example throughout this paper. There are two agents, namely agent $a_1$ starting in location $\start_1=\mathbb A$ at time step 0 and agent $a_2$ starting in location $\start_2=\mathbb C$ at time step 1. There are two stations, namely station $s_1$ with target $\goal_1=\mathbb E$ and station $s_2$ with target $\goal_2=\mathbb F$. We are given time window $[0,6)$ with processing time $T=2$. One optimal solution assigns station $s_2$ and path $\route{\pi_1(0)=\mathbb A, \mathbb B, \mathbb D, \mathbb F=\pi_1(3)}$ to agent $a_1$ and station $s_1$ and path $\route{\pi_2(1)=\mathbb C, \mathbb C, \mathbb D, \mathbb E = \pi_2(4)}$ to agent $a_2$. Both agents occupy working slot $[4,6)$ of their assigned stations, resulting in a total idle time of 8.
	
\subsection{Methodology}
	In this section, we present two efficient algorithms based on a novel min-cost max-flow framework, called the ITO flow framework. The first algorithm uses the ITO flow framework to first assign stations to agents, using their estimated arrival time at every station as path cost, so that the (estimated) total idle time is minimized. It then uses a MAPF algorithm to find collision-free paths for the agents to their assigned stations. The second algorithm combines the ITO flow framework with a MAPF flow framework. The resulting flow framework, called the PITO flow framework, simultaneously assigns stations to and finds collision-free paths for the agents so that the (actual) total idle time is minimized.
	
	
	\subsection{The ITO Flow Framework}
	
\begin{figure}[t]
		\centering
		\includegraphics[width=.9\columnwidth]{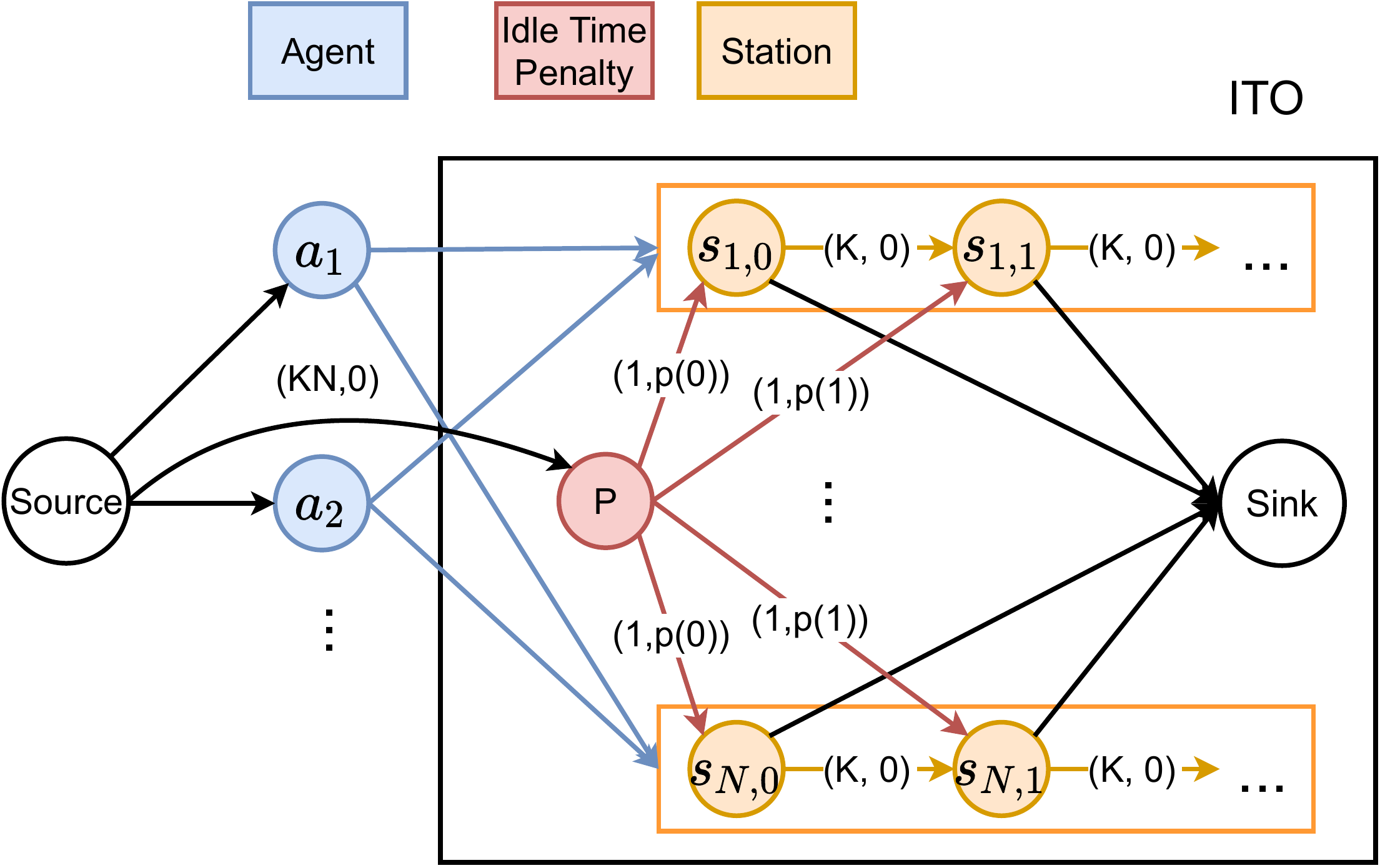}
		\caption{The ITO flow network. Each edge is annotated with (capacity, cost). Unit-capacity and zero-cost edges are not annotated. The red vertex and edges are not used in the unweighted ITO flow network.}
		\label{fig:ITO}
	\end{figure}

\begin{figure}[t]
		\centering
		\includegraphics[width=.85\columnwidth]{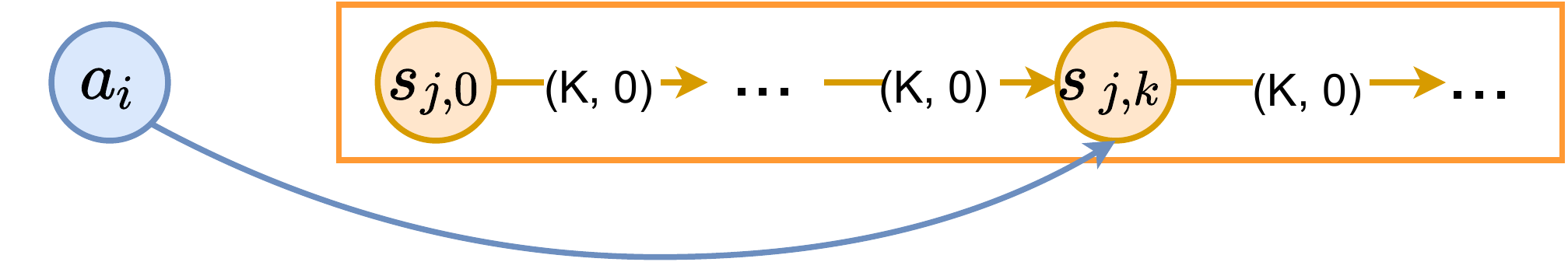}
		\caption{Connection from an agent vertex to a working slot vertex.}
		\label{fig:station_queue}
	\end{figure}

	Given a One-Shot TAPF instance, we construct an unweighted ITO flow network $\mathcal{G=(V,E)}$. Figure \ref{fig:ITO} illustrates the construction. We create an agent vertex (blue vertex) $a_i\in \mathcal V$ for each agent $a_i$. A unit-capacity edge connects the source vertex to each agent vertex. We create a station sequence (orange rectangular substructure) of working slot vertices (orange vertices) $s_{j,k} \in \mathcal V$ for each station $s_j$ and $k=0,\ldots,K-1$. Vertex $s_{j,k}$ represents working slot $[kT,(k+1)T)$ of station $s_j$. We create an edge of unit capacity from agent vertex $a_i$ to working slot vertex $s_{j,k}$ if and only if $k$ is the smallest number such that the arrival time of agent $a_i$ at station $s_j$ is no later than $kT$. An example of such an edge is shown in Figure \ref{fig:station_queue}, which redraws an abstract edge (blue edge) from an agent vertex to a station sequence in Figure \ref{fig:ITO}. Creating edges from agent vertices to working slot vertices requires a method for estimating the arrival times of agents at stations. One such method is to consider individual time-minimal paths for each agent $a_i$ from its start location $\start_i$ to the target $\goal_j$ of each station $s_j$ (while ignoring collisions with other agents).
	We also create an edge of unit capacity from each working slot vertex $s_{j,k}$ to the sink vertex to enforce the constraint that each working slot is occupied by at most one agent because at most one unit of flow can flow through the edge. For each $k=1,\ldots,K-1$ and each station $s_j$, we also create an edge of capacity $K$ from $s_{j,k-1}$ to $s_{j,k}$ to allow for an agent to occupy any working slot later than the earliest working slot that it is allowed to occupy at the same station.
	
	All edges in the unweighted ITO flow network have integral capacities, and therefore an integer max-flow can be found in polynomial time. Such a max-flow corresponds to an assignment of (working slots of) stations to agents that approximately maximizes the number of occupied working slots of all stations in a given time window (since the arrival times of all agents at all stations are estimates). However, this does not mean that all agents are assigned (a working slot of) a station. Agents that are not assigned a station in the max-flow are assigned the NULL station, thereby not affecting the total idle time.
	
	\begin{thm}
	For given arrival times of agents at all stations, a max-flow on the unweighted ITO flow network corresponds to an assignment of stations to agents that minimizes the total idle time within a given time window.
	\end{thm}
		
\begin{figure}[t]
		\centering
		\includegraphics[width=.75\columnwidth]{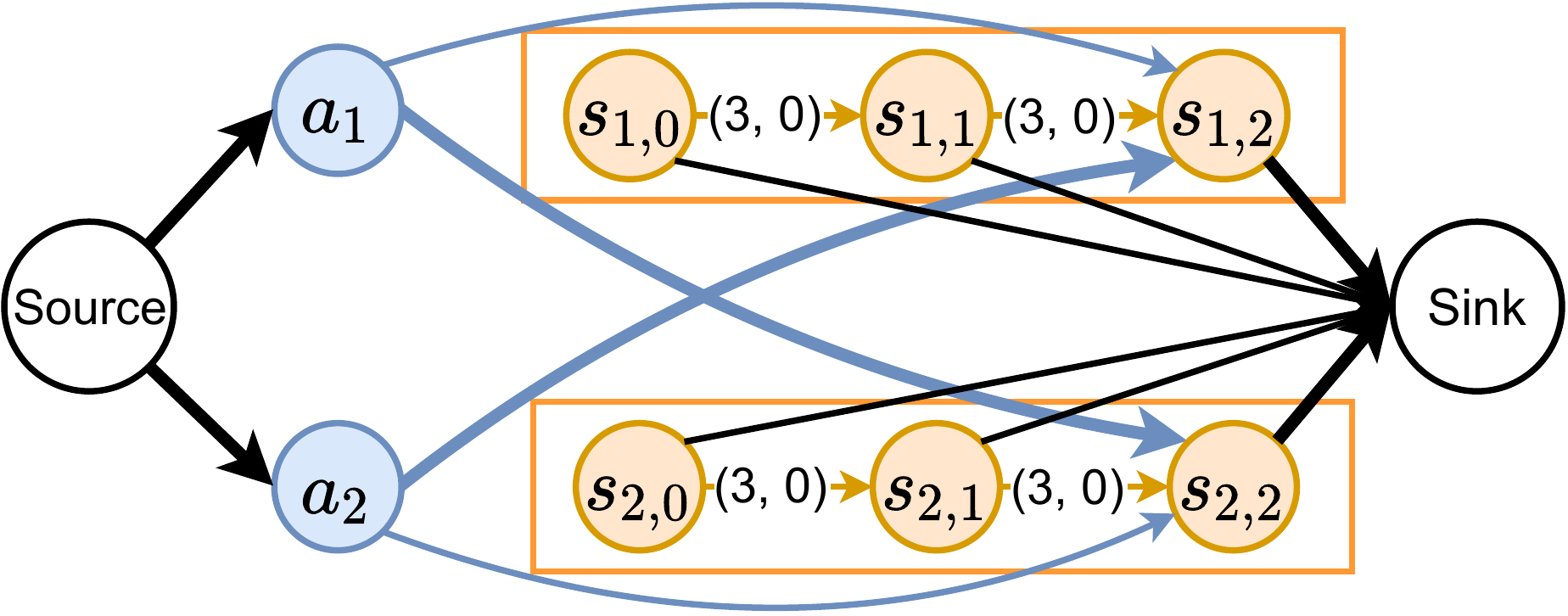}
		\caption{The unweighted ITO flow network for the running example. The estimated arrival times of both agents at either station are 3. There exists a max-flow where the bold blue and bold black edges are saturated.}
		\label{fig:MAPF_ITO}
	\end{figure}

	Figure \ref{fig:MAPF_ITO} shows the unweighted ITO flow network for our running example from Figure \ref{fig:MAPF}.
	
	
	
	The weighted ITO flow framework also uses an idle time penalty vertex $P$ (red vertex), as shown in Figure \ref{fig:ITO}. An edge of capacity $KN$ connects it from the source vertex. Unit-capacity edges connect it to each working slot vertex. An edge from $P$ to working slot vertex $s_{j,k}$ has unit capacity and a positive cost defined by a penalty function $p(k)\geq 0$. All other edges are of zero cost. $P$ and the edges emanating from it serve as a tie-breaking mechanism among multiple max-flow solutions. By using an appropriate penalty function, agents can be biased toward earlier working slots, favoring earlier availability for obtaining parcels at the stations, and making themselves available sooner for new tasks in the context of the lifelong version of TAPF, as described there. Such a mechanism comes at a cost since it requires us to solve a min-cost max-flow problem instead of a max-flow problem.
	
	The ITO flow network does not find paths for the agents to their assigned stations and thus does not avoid collisions, including vertex collisions at targets. Our first algorithm, called \emph{ITO+MAPF}, therefore uses a max-flow algorithm on the ITO flow network in the first phase to assign stations to the agents and any MAPF algorithm in the second phase to find collision-free paths for the agents to their assigned stations. Sortation centers often exhibit well-formedness \cite{CapVK15} that results in solvable MAPF instances.
	
	
	\subsection{The PITO Flow Framework}
	\begin{figure}[t]
		\centering
		\includegraphics[width=1\columnwidth]{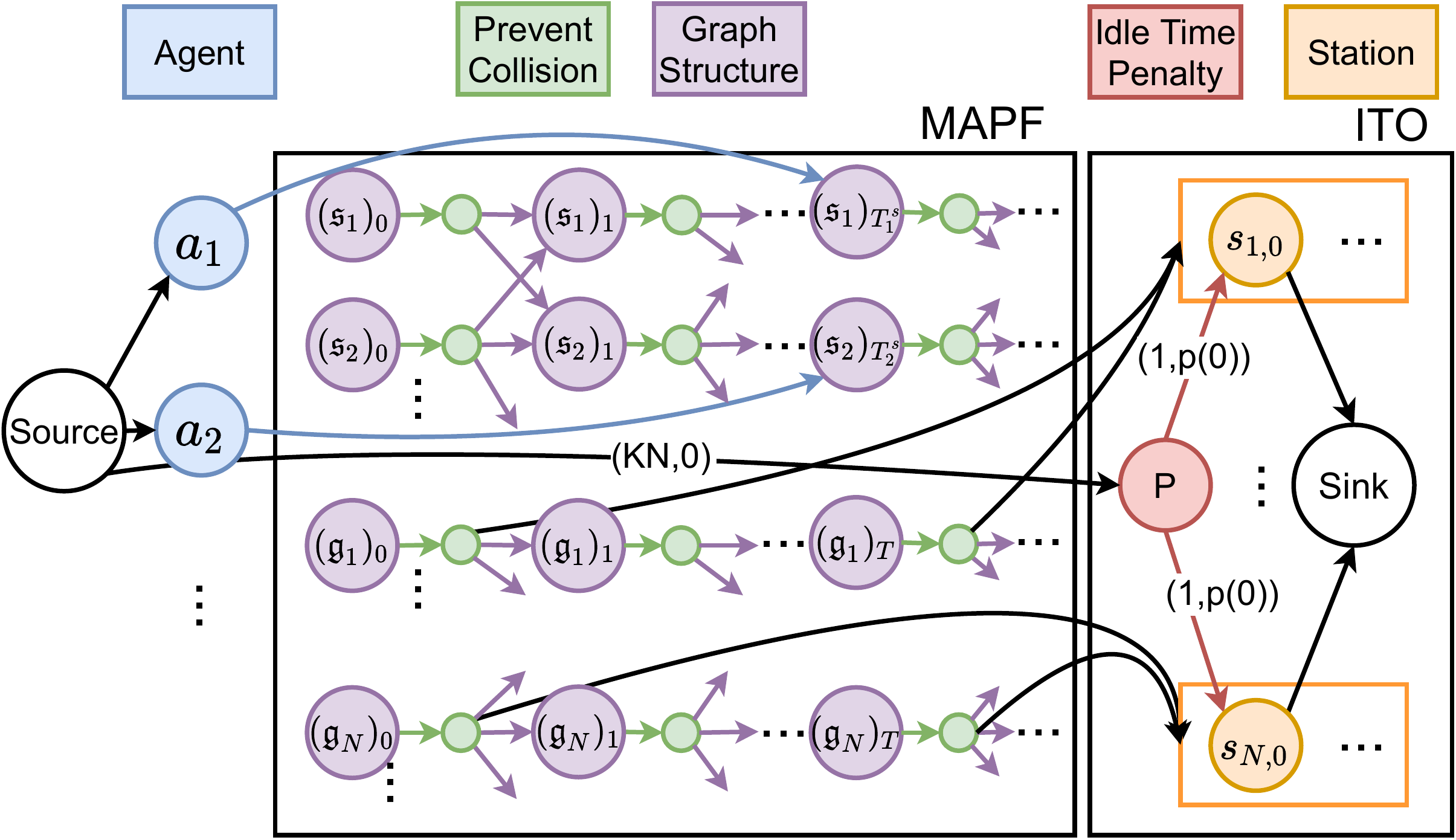}
		\caption{The PITO flow network. The MAPF component on the left represents the Anonymous MAPF flow network. The ITO component on the right represents the ITO flow network, as described in Figures~\ref{fig:ITO} and \ref{fig:station_queue}. Each edge is annotated with (capacity, cost). Unit-capacity and zero-cost edges are not annotated. The red vertex and edges are not used in the unweighted PITO flow network.}
		\label{fig:PITO}
	\end{figure}
	
	ITO+MAPF solves One-Shot TAPF by designing principled algorithmic techniques for its two sub-problems, namely station assignment and path finding. However, it depends on estimates of the arrival times of the agents at the stations. In order to remove this dependency, we develop a new framework, called the PITO flow network, that exploits the fact that station assignment and MAPF can both be tackled with flow methods. It thus combines them in a single framework by combining the ITO flow network with the Anonymous MAPF flow network~\cite{YuLav13STAR}.
	
	
	Given a One-Shot TAPF instance, we construct an unweighted PITO flow network $\mathcal{G= (V,E)}$. Figure~\ref{fig:PITO} illustrates the construction.
	The MAPF component is constructed as follows.
	For each location $u\in V$ and each time step $t=0,\ldots,(K-1)T$, we create a location vertex $u_t \in \mathcal V$ (purple vertex) and an auxiliary vertex $u_t' \in \mathcal V$ (green vertex). We create an edge of unit capacity from $u_t$ to $u_t'$ to prevent vertex collisions in location $u$ at time step $t$. For each edge $(u,v)\in E$ and each time step $t=0,\ldots,(K-1)T-1$, we create two unit-capacity edges, one from $u_t'$ to $v_{t+1}$ and one from $v_t'$ to $u_{t+1}$, to allow agents to traverse edge $(u,v)$ between time steps $t$ and $t+1$. In Anonymous MAPF, edge collisions do not need to be prevented since they can be avoided during post processing by swapping the identities of the two colliding agents and letting them wait in their current locations for one time step. We also create an edge of unit capacity from each agent vertex $a_i$ to location vertex $(\start_i)_{\startt i}$ to let agent $a_i$ enter the environment at its start time $\startt i$ in its start location $\start_i$. All edges in the MAPF component are of zero cost.
	The ITO component is constructed as follows: It resembles the ITO flow network, except for the edges from agent vertices to working slot vertices. We create an edge of unit capacity from auxiliary vertex $(\goal_j)_{kT}'$ to each working slot vertex $s_{j, k}$ to allow one agent in location $\goal_j$ at time step $kT$ to occupy working slot $[kT,(k+1)T)$ of station $s_j$. All these edges are of zero cost. The MAPF flow network is sufficiently large to allow agents to find paths to all stations within the given time window, if possible. The unweighted PITO flow network, like the unweighted ITO flow network, does not include the idle time penalty vertex $P$ and its incident edges in the ITO component, but the weighted PITO flow network does. The role of $P$ in the unweighted PITO flow network is analogous to that in the unweighted ITO flow network.
	
	All edges in the unweighted PITO flow network have integral capacities, and therefore an integer max-flow can be found in polynomial time. Such a max-flow corresponds to an assignment of (working slots of) stations to agents that maximizes the number of occupied working slots of all stations in a given time window and also provides collision-free paths for the agents to their assigned stations. As in the ITO flow network, agents that are not assigned a station in the max-flow are assigned the NULL station (and are thus removed from the environment), thereby not affecting the total idle time.
	
	\begin{thm}
	A max-flow on the unweighted PITO flow network corresponds to an assignment of stations to agents and collision-free paths for all active agents to their assigned stations that together minimize the total idle time within a given time window.
	\end{thm}
	
	
	\begin{figure}[t]
		\centering
		\includegraphics[width=.7\columnwidth]{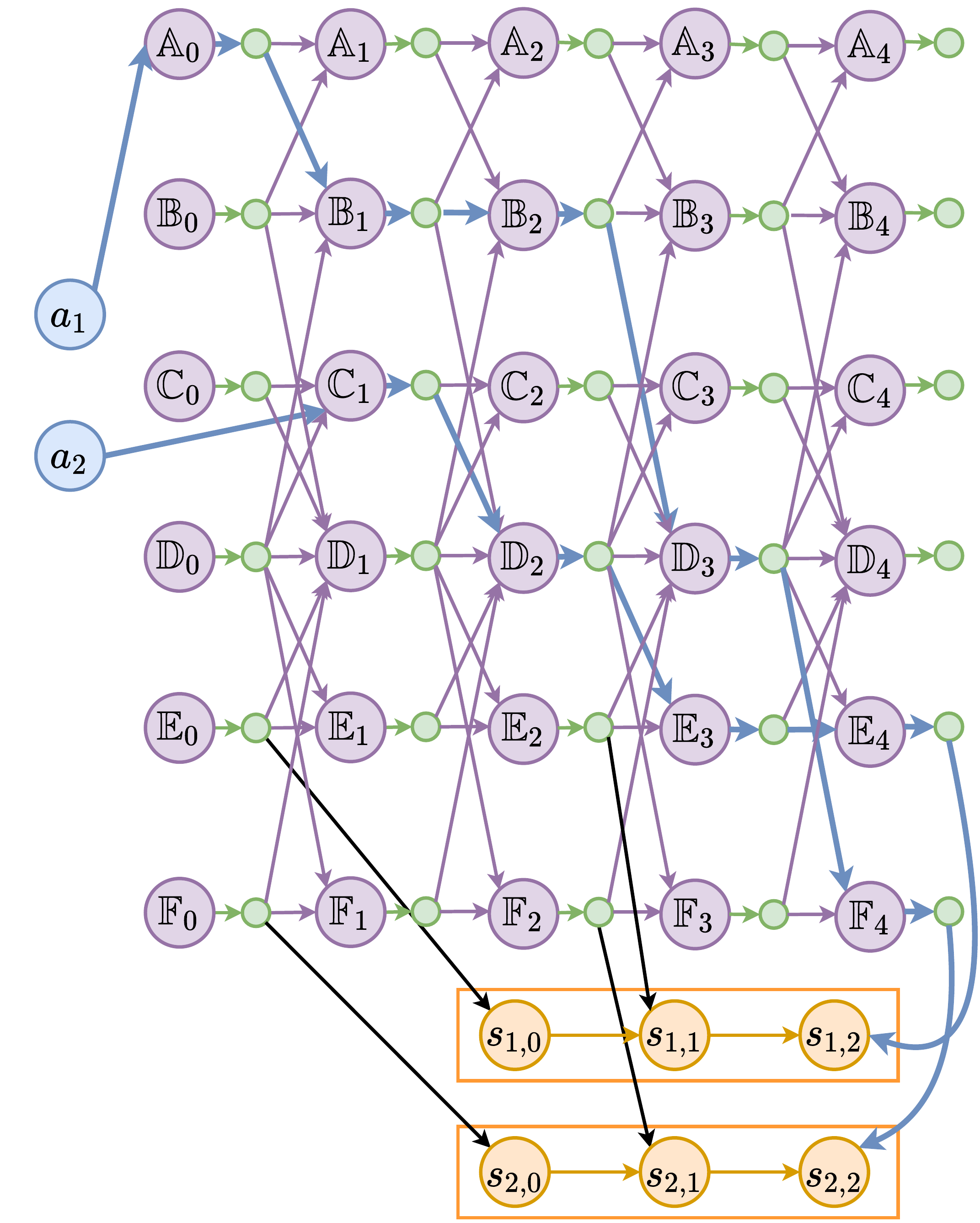}
		\caption{The unweighted PITO flow network for the running example. The source and sink vertices are not shown. There exists a max-flow where the bold blue edges are saturated.}
		\label{fig:MAPF_PITO}
	\end{figure}
	
	Figure \ref{fig:MAPF_PITO} shows the unweighted PITO flow network for our running example from Figure \ref{fig:MAPF}.
	
	\section{Lifelong TAPF}\label{sec:lifelong}
	
	In this section, we consider the lifelong version of TAPF, \emph{Lifelong TAPF}.
	We are interested in minimizing the total idle time and thus maximizing the number of parcels obtained by the agents, which directly relates to maximizing the throughput of the sortation center. Because the horizon of its operating time is not known a priori, a good strategy for solving Lifelong TAPF is to repeatedly solve One-Shot TAPF instances such that a solution for one instance facilitates an effective solution for the next instance. Therefore, methods for solving Lifelong TAPF can be built on methods for solving One-Shot TAPF. For example, a One-Shot TAPF instance could be solved every $W\leq KT$ time steps. When a One-Shot TAPF instance with a fixed time window $[0,KT)$ is solved, both agents moving to their assigned stations and agents delivering parcels that can occupy a working slot after the delivery and within the time window should be included. The other agents can be ignored because they do not affect the total idle time. In reality, they could be treated as moving obstacles by our algorithms.
	
	We develop algorithms for Lifelong TAPF based on our two One-Shot TAPF algorithms that use the ITO flow network and the PITO flow network, respectively. We refer to these algorithms as ITO-L and PITO-L, where L indicates the lifelong version. Both ITO-L and PITO-L exploit the idle time penalty vertex $P$. The edges emanating from $P$ are designed to push a unit of flow through any working slot vertex not occupied by any agent. Since the cost of this edge is defined by the positive penalty function $p(k)$, an appropriate choice of $p(k)$ can bias agents toward occupying earlier or later working slots within the given time window. For example, if $p(k)$ monotonically decreases with $k$, it biases agents toward occupying earlier working slots. An exponentially decreasing function, such as $p(k)=N^{-k}$, can even force the agents to occupy the earliest possible working slots. The unweighted ITO and PITO flow networks result for the constant function $p(k)$. In the experiments, we use the linear function $p(k)=K-k$, which experimentally resulted in a similar solution quality as the exponentially decreasing function $p(k)=N^{-k}$ in our test trials but avoids numerical underflows for a large $k$.
	
	A good baseline strategy to compare ITO-L and PITO-L against is to call the Hungarian method \cite{Kuhn1955} at regular time intervals. For each One-Shot TAPF instance, the Hungarian method, labeled as H(Inf), computes an assignment of stations to all agents that minimizes the sum of the estimated arrival times of the agents at their assigned stations. A MAPF algorithm then computes collision-free paths for the agents to their assigned stations. H(Inf) results in a greedy policy that assigns each agent a station with the smallest estimated arrival time. This assignment can result in unbalanced station queues if agents are located in the vicinity of the same station.	
	To address this issue, we develop an extension of the Hungarian method, labeled as H($Q$) for $Q>0$, that runs the Hungarian method $\lceil\frac{M}{NQ}\rceil$ times until all agents are assigned a station. In each but the last iteration, H($Q$) computes an assignment of stations to $NQ$ agents that have not yet been assigned stations, where each station is assigned to $Q$ agents, so that the sum of the estimated arrival times of the agents at their assigned stations is minimized. In the last iteration, H($Q$) computes an assignment of stations to all agents that have not yet been assigned stations, where each station is assigned to at most $Q$ agents, so that the sum of the estimated arrival times of the agents at their assigned stations is minimized.

	
	\section{Experiments}
In this section, we compare different algorithms for Lifelong TAPF. We implement all algorithms in Java and conduct the experiments on a CentOS 6.9 server with an Intel Xeon E5-2682@2.5GHz processor and 256GB of memory.

We convert the H(Inf), H($Q$) [H($Q$) for $Q>1$], and H(1) [H($Q$) for $Q=1$] baseline algorithms for One-Shot TAPF to their lifelong versions, H(Inf)-L, H($Q$)-L, and H(1)-L, respectively, by using a MAPF algorithm after the process of assigning stations to agents. We use Prioritized-Based Search (PBS)~\cite{ma2019PBS} because sortation centers often exhibit well-formedness \cite{CapVK15}, for which backtrack-free PBS is possible. We have to modify it in two ways to make it apply to One-Shot TAPF. First, agents can enter the environment at non-zero start time steps. Second, agents are removed from the environment once they occupy a working slot.

We use a Primal-Dual algorithm to solve the min-cost max-flow problems for ITO-L and PITO-L. 




\subsection{Agent Simulator}

\begin{figure}[t]
	\centering
	\begin{subfigure}[t]{.5\columnwidth}
		\centering
		\label{fig:sortation_map1}
		\includegraphics[width=0.915\columnwidth]{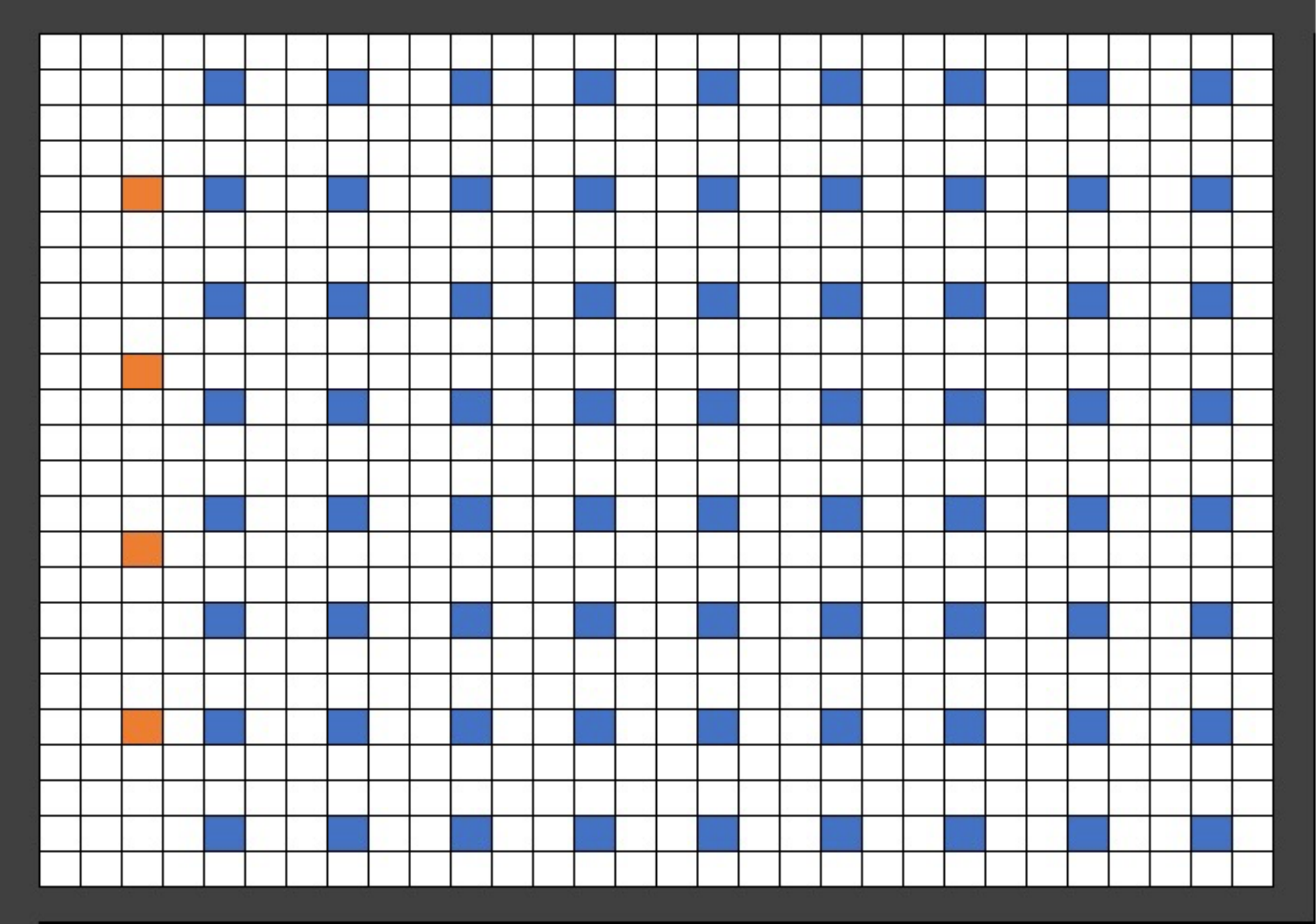}
		\caption{$24\times30$.}
	\end{subfigure}%
	\begin{subfigure}[t]{.5\columnwidth}
		\centering		
		\label{fig:sortation_map2}
		\includegraphics[width=\columnwidth]{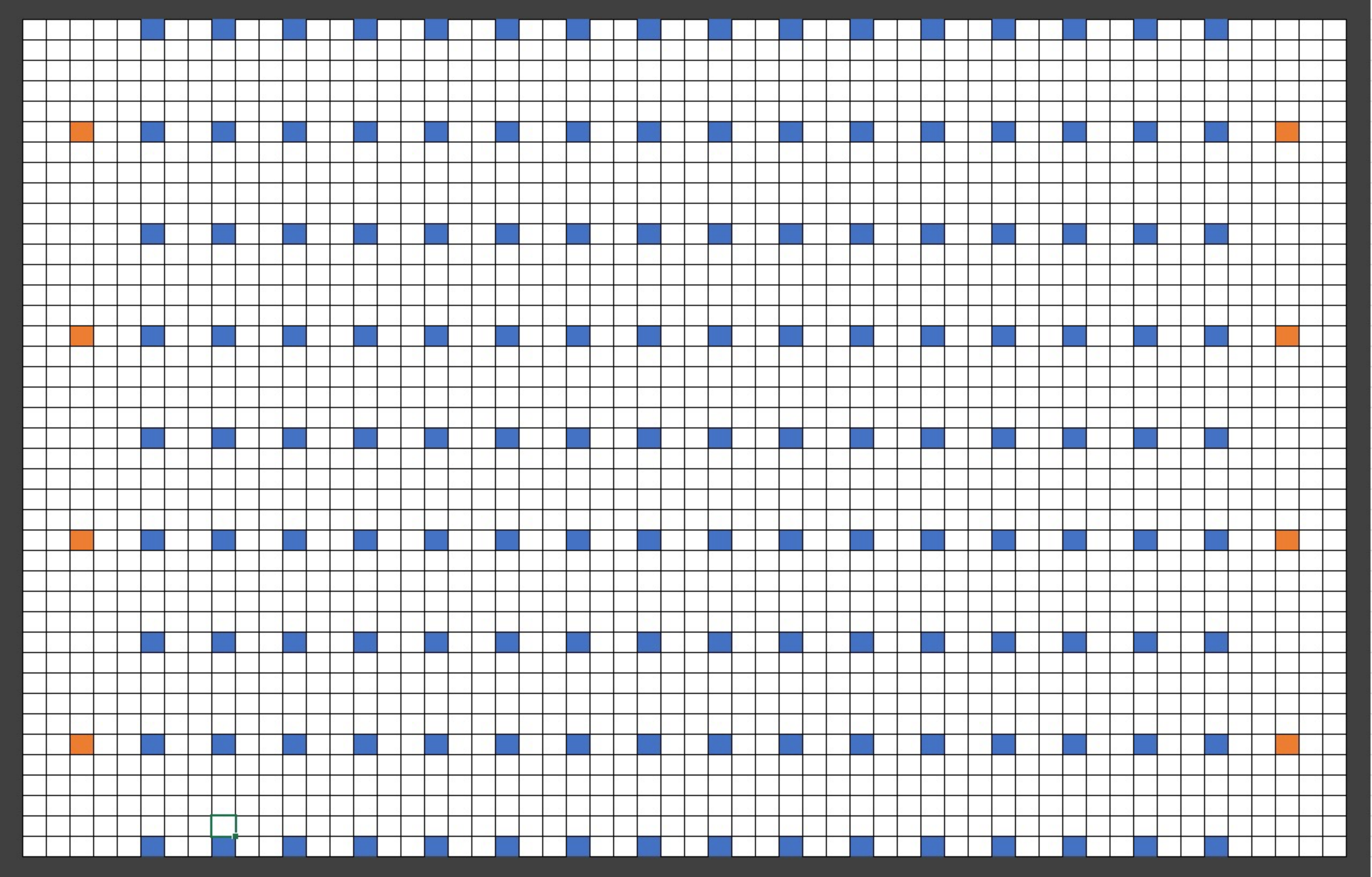}
		\caption{$41\times56$.}
	\end{subfigure}
	\caption{Maps of simulated sortation centers.}
	\label{fig:sortation_map}
\end{figure}

Our first set of experiments is done on an agent simulator with the two sortation centers shown in Figure~\ref{fig:sortation_map}. The first sortation center has $24\times30$ cells with 4 stations on the left side and 72 sorting bins distributed uniformly to the right of them. The second sortation center has $41\times56$ cells with 4 stations on each side and 144 sorting bins distributed uniformly in between. The agent simulator starts at time step 0 (with all agents in a random sorting bin cell) and ends at time step 600, with $\frac {600} W$ strides of $W$ time steps each within which a solution to each One-Shot TAPF instance is computed with a sufficiently large time window so that no agent is assigned the NULL station by any algorithm. For our agent simulator, we use processing time $T=10$ and a sufficiently large $K=9$. Each One-Shot TAPF instance considers only those agents whose start time steps are before the end of the stride. The agent simulator does not model the paths of agents to the sorting bins of their parcels. It thus makes the simplifying assumption that an agent occupying working slot $[kT,(k+1)T)$ is removed from the environment at time step $kT+1$ and added to the environment at time step $kT+1+\kappa$ in a random sorting bin cell, where $\kappa$ represents the average delivery time of a parcel. For our agent simulator, we use $\kappa=30$. We use $Q=\frac{M}{N} + 5$ for H($Q$). 

\begin{figure}[t]
	\centering
	\begin{subfigure}[b]{.5\columnwidth}
		\centering
		\label{fig:idleTime_smallMa}
		\includegraphics[width=\columnwidth]{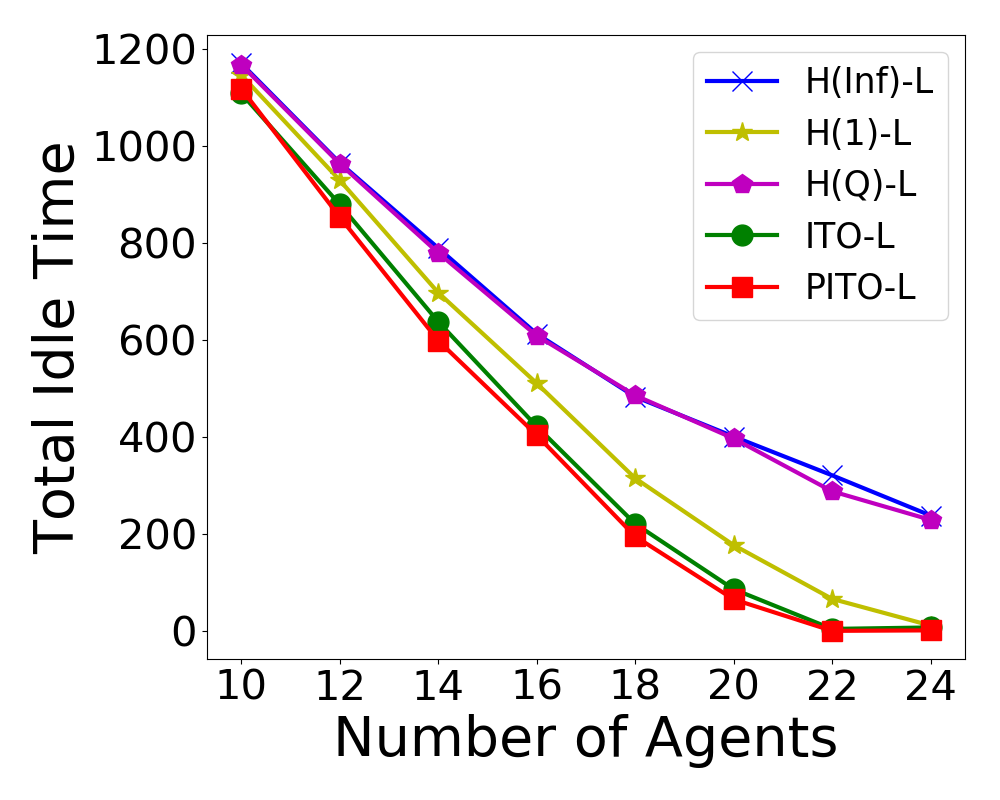}
		\caption{$24\times30$, $W=30$.}
	\end{subfigure}%
	\begin{subfigure}[b]{.5\columnwidth}
		\centering		
		\label{fig:idleTime_bigMap}
		\includegraphics[width=\columnwidth]{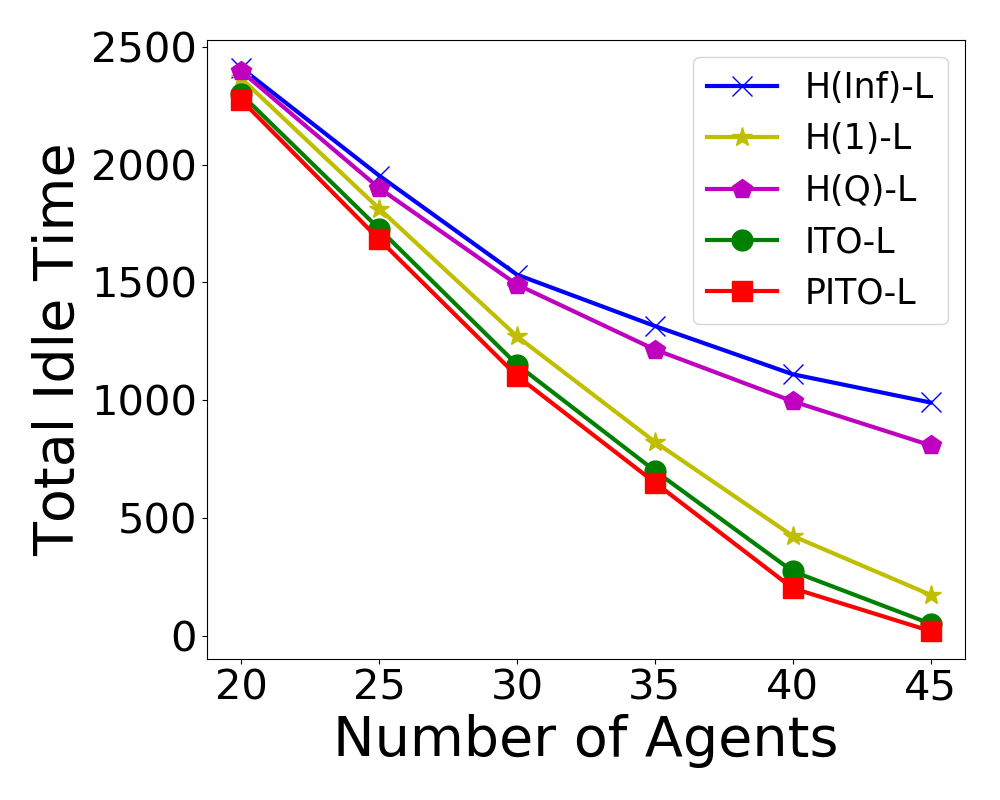}
		\caption{$41\times56$, $W=30$.}
	\end{subfigure}
	\caption{Total idle times for varying numbers of agents.}
	\label{fig:idle_time}
\end{figure}

\begin{figure}[t]
	\centering
	\begin{subfigure}[b]{.5\columnwidth}
		\centering
		\label{fig:comp_time_small}
		\includegraphics[width=\columnwidth]{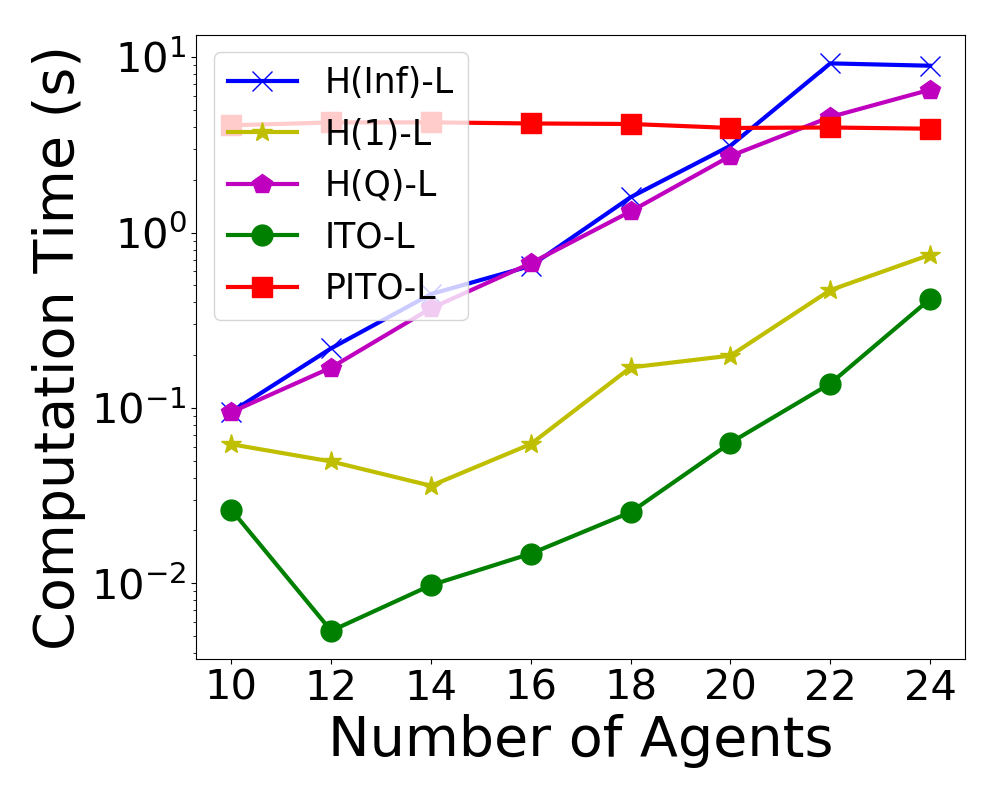}
		\caption{$24\times30$, 20 agents, $W=30$.}
	\end{subfigure}%
	\begin{subfigure}[b]{.5\columnwidth}
		\centering		
		\label{fig:comp_time_big}
		\includegraphics[width=\columnwidth]{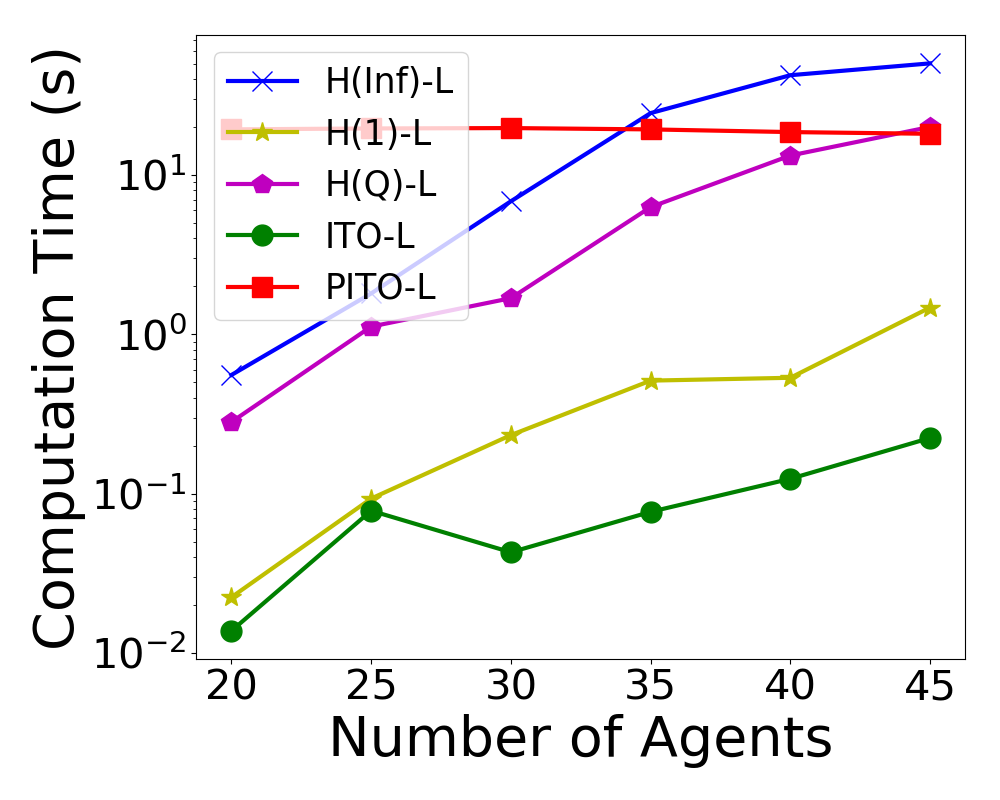}
		\caption{$41\times56$, 40 agents, $W=30$.}
	\end{subfigure}
	\caption{Average computation times for solving each One-Shot TAPF instance for varying numbers of agents.}
	\label{fig:comp_time}
\end{figure}

\begin{figure}[t]
	\centering
	\begin{subfigure}[b]{.5\columnwidth}
		\centering
		\label{fig:finished_tasks_small}
		\includegraphics[width=\columnwidth]{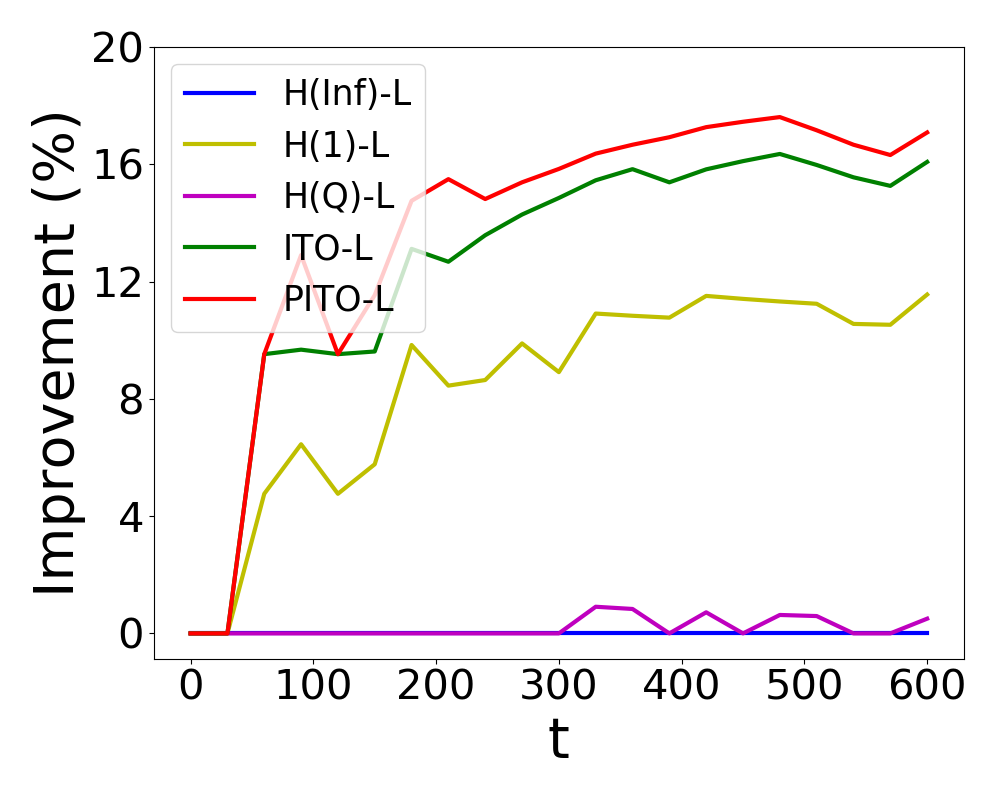}
		\caption{$24\times30$, 20 agents, $W=30$.}
	\end{subfigure}%
	\begin{subfigure}[b]{.5\columnwidth}
		\centering		
		\label{fig:finished_tasks_big}
		\includegraphics[width=\columnwidth]{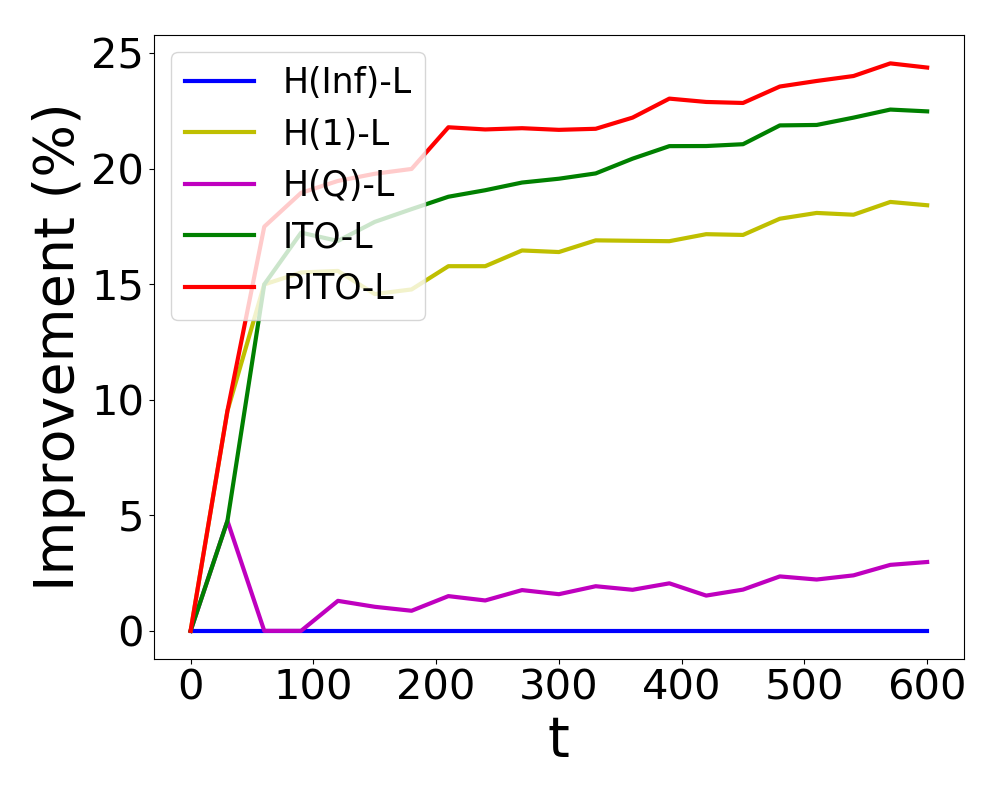}
		\caption{$41\times56$, 40 agents, $W=30$.}
	\end{subfigure}
	\caption{Percentage improvements over H(Inf)-L with respect to the total number of parcels obtained by time step $t$.}
	\label{fig:finished_task}
\end{figure}

\noindent\textbf{Experiment 1:} Figures \ref{fig:idle_time}(a)\&(b) compare the total idle times for varying numbers of agents in the two sortation centers. As expected, for a small number of agents, the differences between the algorithms are small. In general, as the number of agents increases, the differences between the algorithms increase. However, PITO-L, ITO-L, and H(1)-L perform well for all numbers of agents. PITO-L outperforms the other algorithms, while ITO-L is a close competitor. Our modification of the Hungarian method H(1)-L also performs well.


\noindent\textbf{Experiment 2:} Figures \ref{fig:comp_time}(a)\&(b) compare the average computation times on a logarithmic scale for solving each One-Shot TAPF instance for varying numbers of agents. ITO-L outperforms the other algorithms, while H(1)-L is second-best. The target-assignment algorithms H(Inf), H($Q$), H(1), and ITO run in polynomial time for each One-Shot TAPF instance and very fast in our experiments. However, the subsequent path-planning algorithm PBS can run in exponential time in the number of agents and slowly \cite{ma2019PBS}. It runs faster for H(1) and ITO since they produce more balanced assignments of agents to stations and thus also potentially fewer collisions of the resulting individual paths than H(Inf) and H($Q$). The target-assignment and path-planning algorithm PITO-L always runs in polynomial time for each One-Shot TAPF instance. Its computation time remains unchanged as the number of agents increases since the size of the PITO flow network does not increase. It tends to run slowly in our experiments due to the large size of the network but could potentially be sped up with a better min-cost max-flow solver.

\noindent\textbf{Experiment 3:} Figures \ref{fig:finished_task}(a)\&(b) compare the percentage improvements over H(Inf)-L with respect to the total number of parcels obtained by time step $t$. The ranking of the algorithms is similar to that in Figure \ref{fig:idle_time}. The better performances of PITO-L and ITO-L are expected to translate to better throughputs in actual sortation centers, where even a 10\% improvement is significant.


\begin{figure}[t]
	\centering
	\begin{subfigure}[b]{.5\columnwidth}
		\centering
		\label{fig:interval_small}
		\includegraphics[width=\columnwidth]{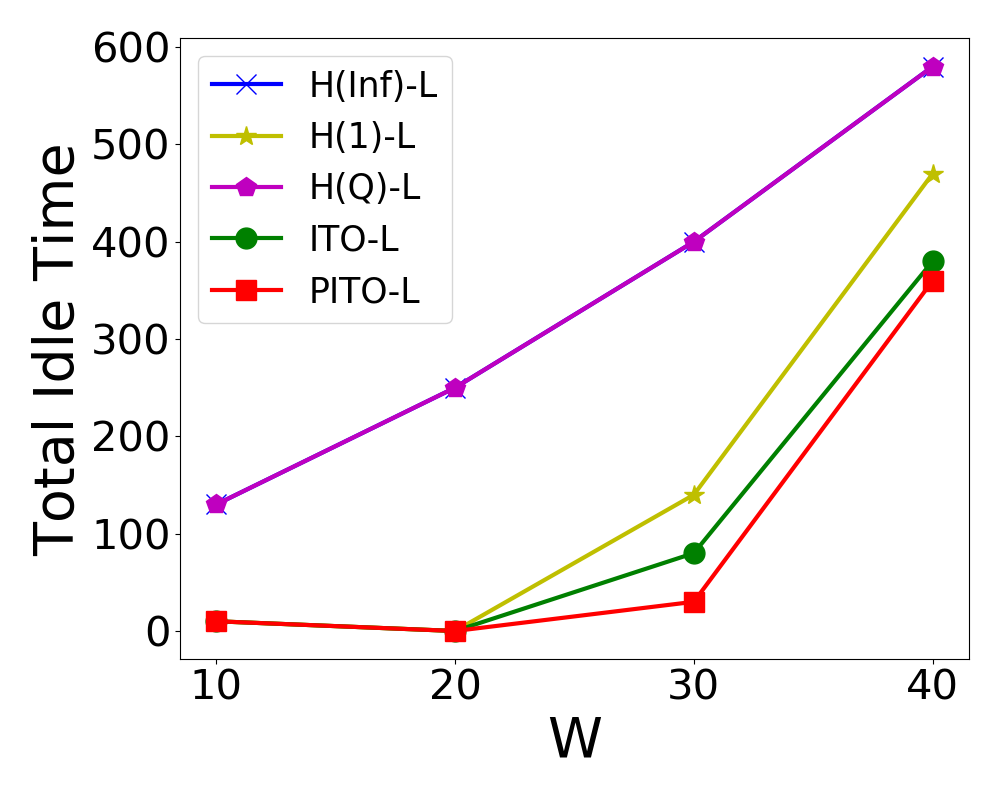}
		\caption{$24\times30$, 20 agents.}
	\end{subfigure}%
	\begin{subfigure}[b]{.5\columnwidth}
		\centering		
		\label{fig:interval_big}
		\includegraphics[width=\columnwidth]{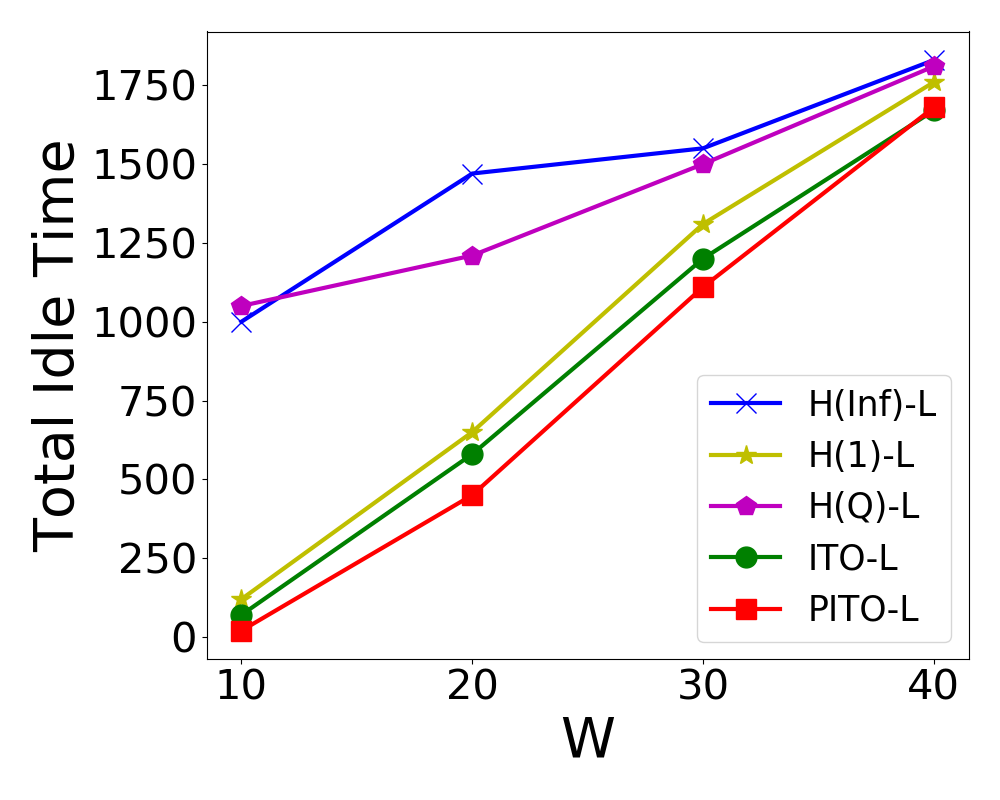}
		\caption{$41\times56$, 40 agents.}
	\end{subfigure}
	\caption{Total idle times for varying values of $W$.}
	\label{fig:interval}
\end{figure}

\noindent\textbf{Experiment 4:} Figures \ref{fig:interval}(a)\&(b) compare the total idle times for varying values of $W$. Again, the ranking of the algorithms is similar to that in Figure \ref{fig:idle_time}. Smaller values of $W$ result in smaller total idle times since the agents adjust more frequently to the system dynamics.

\subsection{Industrial Simulator}

While PITO is optimal for One-Shot TAPF if no agent is assigned the NULL station, it requires the agents to safely execute the computed paths. In reality, computing kinodynamically feasible paths for all agents is hard. A polynomial-time post-processing step, called MAPF-POST \cite{HoenigICAPS16}, could be used to reinstate some of the kinodynamic constraints, such as the maximum velocities of agents. It is thus applicable to sortation centers and other multi-agent systems, such as Amazon fulfillment centers \cite{kiva}, that use simple agents without complex kinodynamic constraints. On the other hand, since ITO and H($Q$) do not compute actual paths for the agents and can be combined with off-the-shelf motion-planning algorithms, they are applicable even in the presence of complex kinodynamic constraints.



We thus implement ITO-L and H($Q$)-L on a real-time industrial simulator that includes higher-order dynamic constraints. The simulator uses data collected from real sortation centers, including about parcel distribution, human efficiency, machine delay, and exceptions and uses a physical model of custom-made parcel sorting robots from Guozi Robotics (http://www.smartlogisticsolutions.com/) that have a rectangular shape of \SI{0.45x0.37}{m}, a maximum translational velocity of \SI{2.5}{m/s}, a maximum translational acceleration/deceleration of \SI{1}{m/s^2}, a maximum rotational velocity of \SI{5}{rad/s}, and a maximum rotational acceleration/deceleration of \SI{8}{rad/s^2}. A robot controller receives commands and reports the status of all robots every \SI{100}{ms}, including their locations, orientations, velocities, and errors. Path planning is done centralized with A* for motion planning and a mechanism for deadlock avoidance. The simulator uses strides of two seconds within which a solution to each One-Shot TAPF instance is computed. The computation of the estimated arrival times of all robots at each station is parallelized based on the costs of their individually time-minimal paths, the current traffic, and historical statistics.



\begin{figure}[t]
	\centering
	\begin{subfigure}[t]{.54\columnwidth}
		\centering
		\includegraphics[width=\columnwidth]{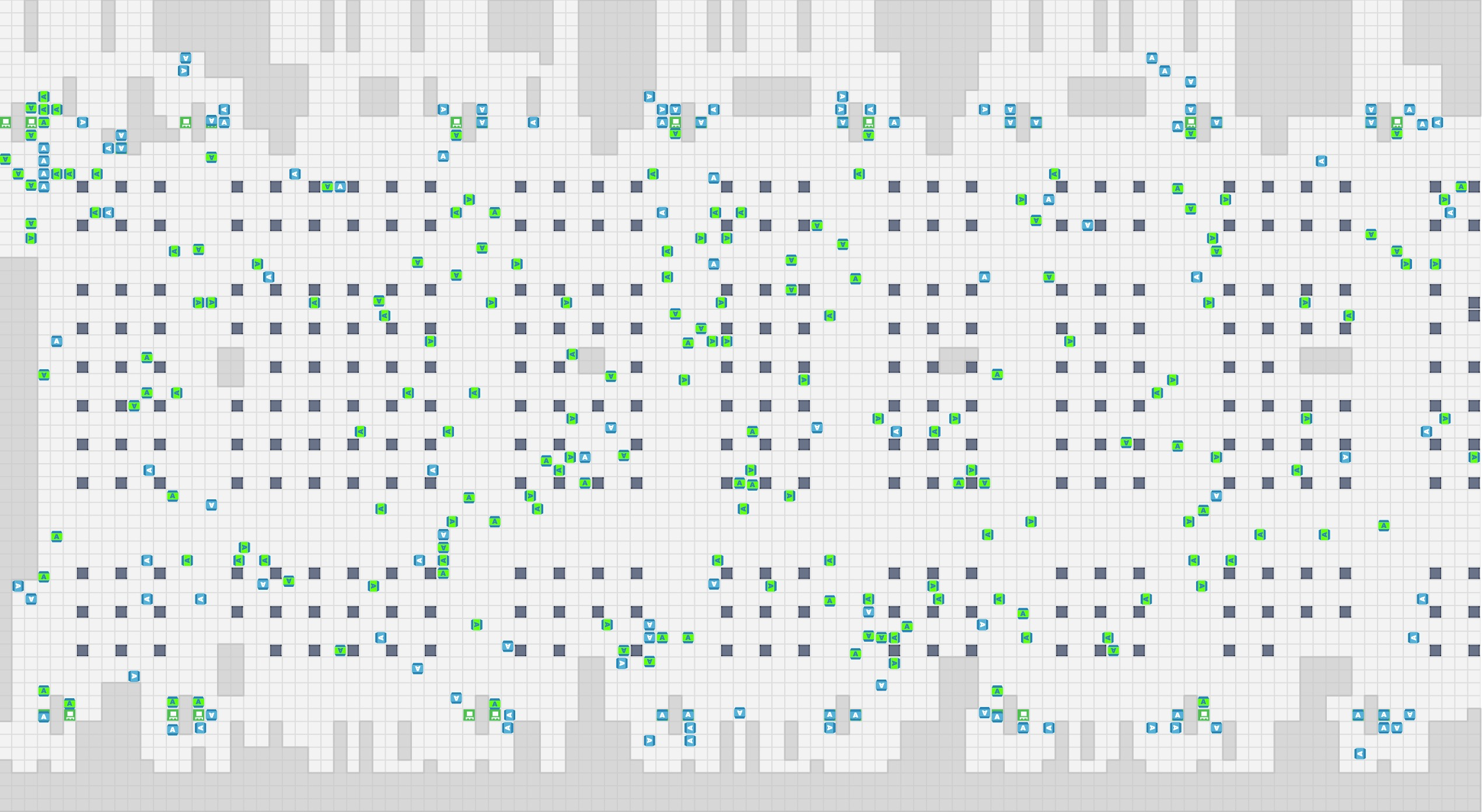}
		\caption{Industrial simulator executing ITO-L.}
		\label{fig:simulator_screenshot}
	\end{subfigure}\hfill
	\begin{subfigure}[t]{0.45\columnwidth}
		\centering
		\includegraphics[width=\columnwidth]{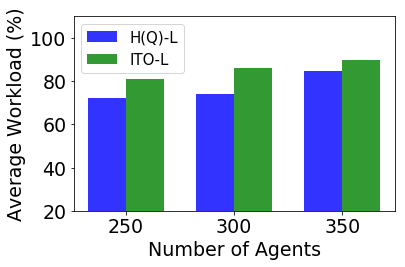}
		\caption{Average workload for varying numbers of agents.}
		\label{fig:simulator_result}
	\end{subfigure}
	\caption{Evaluation of ITO-L and H($Q$)-L.}
	\label{fig:simulator}
\end{figure}

Figure~\ref{fig:simulator_screenshot} shows a screen shot of the industrial simulator executing ITO-L on a sortation center with $63 \times 89$ cells of size \SI{0.5x0.5}{m} each and 33 stations. We use $T=\SI{4}{s}$ and a sufficiently large $K=75$. We use $Q=15$. Figure~\ref{fig:simulator_result} compares the average workload percentage for varying numbers of agents. The average workload percentage is the percentage of time that stations are occupied, averaged over all stations. ITO-L outperforms H($Q$)-L by $9.01\%$, $11.97\%$, and $4.86\%$ for $250$, $300$, and $350$ agents, respectively, and thus has the potential to achieve high throughputs in actual sortation centers. 

\section{Conclusions}

We studied the one-shot and lifelong versions of TAPF in automated sortation centers. To optimize throughput in such centers, we focused on the problem of minimizing the total idle time of the sorting stations. We first presented efficient algorithms based on a novel min-cost max-flow formulation that minimizes the total idle time of all stations in One-Shot TAPF. We then extended these algorithms to Lifelong TAPF. Experimentally, we showed that our algorithms for Lifelong TAPF are efficient and effective, including for up to 350 agents on an industrial simulator.

\section{Acknowledgments}

The research at the University of Southern California was supported by the National Science Foundation (NSF) under grant numbers 1409987, 1724392, 1817189, and 1837779. Hang Ma was supported by Cainiao Smart Logistics Network.

	\small
	\bibliography{references}
	\bibliographystyle{aaai}
\end{document}